\documentclass[letterpaper]{article} 
\usepackage{aaai23}  
\usepackage{times}  
\usepackage{helvet}  
\usepackage{courier}  
\usepackage[hyphens]{url}  
\usepackage{graphicx} 
\usepackage{natbib}  
\usepackage{caption} 
\usepackage{algorithm}
\usepackage{algorithmic}
\usepackage{multirow}
\usepackage{makecell}
\usepackage{bm}
\usepackage{amsmath}
\usepackage{amssymb}
\usepackage{arydshln}

\usepackage{newfloat}
\usepackage{listings}
\DeclareCaptionStyle{ruled}{labelfont=normalfont,labelsep=colon,strut=off} 
\floatstyle{ruled}
\newfloat{listing}{tb}{lst}{}
\floatname{listing}{Listing}
\title{Linking Sketch Patches by Learning Synonymous Proximity \\for Graphic Sketch Representation}
\author{
    Sicong Zang,
    Shikui Tu,
    Lei Xu
}
\affiliations{
    Department of Computer Science and Engineering, Shanghai Jiao Tong University, Shanghai, China\\
    \{sczang, tushikui, leixu\}@sjtu.edu.cn
}

\usepackage{bibentry}

\begin{document}

\maketitle

\begin{abstract}
Graphic sketch representations are effective for representing sketches. Existing methods take the patches cropped from sketches as the graph nodes, and construct the edges based on sketch's drawing order or Euclidean distances on the canvas. However, the drawing order of a sketch may not be unique, while the patches from semantically related parts of a sketch may be far away from each other on the canvas. In this paper, we propose an order-invariant, semantics-aware method for graphic sketch representations. The cropped sketch patches are linked according to their global semantics or local geometric shapes, namely the synonymous proximity, by computing the cosine similarity between the captured patch embeddings. Such constructed edges are \emph{learnable} to adapt to the variation of sketch drawings, which enable the message passing among synonymous patches. Aggregating the messages from synonymous patches by graph convolutional networks plays a role of denoising, which is beneficial to produce robust patch embeddings and accurate sketch representations. Furthermore, we enforce a clustering constraint over the embeddings jointly with the network learning. The synonymous patches are self-organized as compact clusters, and their embeddings are guided to move towards their assigned cluster centroids. It raises the accuracy of the computed synonymous proximity. Experimental results show that our method significantly improves the performance on both controllable sketch synthesis and sketch healing.
\end{abstract}

\section{Introduction}

Free-hand sketches is a traditional medium for social communication and conveying human emotions. They are vivid and impressive, but are always abstract, iconic and lack-of-details. Though sketches are with various visual appearances, a sketch only consists of several pen strokes. It is challenging to learn accurate and robust sketch representations.


Recently, graphic sketch representation is effective for representing sketches. A sketch is cropped into small patches \citep{su2020sketchhealer,qi2022generative} or constructed as coordinates on lattice \citep{qi2021sketchlattice}, regarded as the graph nodes. These nodes are linked by edges according to the Euclidean distances on the canvas (spatial proximity) \citep{qi2021sketchlattice} or the sketch drawing order (temporal proximity) \citep{su2020sketchhealer,qi2022generative}. Usually, graph convolutional network (GCN) \citep{kipf2016semi} is utilized to aggregate the feature from the node itself and the ones from its neighboring nodes to get the final node representation. Such a GCN-based learning is promising on sketch generation as the neighboring nodes are helpful additional sources for the structural patterns, but its performance largely depends on the principle to connect small patches for message passing.

Neither the temporal nor the spatial proximity is robust against the variation of sketch drawings for patch linking. On one hand, the drawing order of a sketch is not unique. For example, when drawing a cat, its head may be drawn before or after its body, which produces different groups of edges in the constructed graph. On the other hand, the patches from semantically related parts of a sketch may be far away from each other on the canvas. For example, the patches of two ears of a cat may not be linked by an edge. The message from a patch of one ear is difficult to reach a patch of the other ear, as it has to go through a long, multi-hop path (if any). The messages would be diluted or interrupted by other patches from different sketch parts along the path.

\begin{figure}[!t]
  	\centering
  	\includegraphics[width=0.9\linewidth]{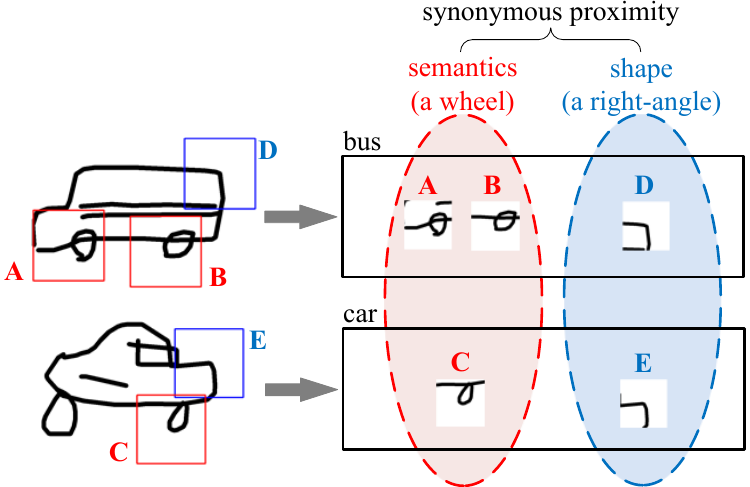}
  	\caption{The definition of synonymous proximity. Sketch patches can be semantically related (A, B and C all contain a wheel) or collect pen strokes with analogous shape (D and E are similar in a right-angle-shape). We summarize these neighboring rules as the synonymous proximity.}
	\label{fig:definition of icon synonym}
\end{figure}

It motivates us to link the patches by semantics or geometric shapes. More specifically, if two cropped sketch patches are semantically related (e.g., the patches A, B and C in Fig. \ref{fig:definition of icon synonym} all contain a wheel), or their collections of pen strokes share analogous shapes (e.g., D and E in Fig. \ref{fig:definition of icon synonym} are similar in a right-angle-shape), we link them by a graph edge and name this neighboring rule as the \emph{synonymous proximity}. Synonym is a term borrowed from the field of linguistics, describing that two words or phrases are with exactly or nearly the same meanings. We use this word to represent the similarity between the content of patches. Compared with the edges linked by temporal or spatial proximity, our principle is \emph{learnable} to adapt to the variation of sketch drawings. Aggregating the messages from the synonymous patches plays a role of denoising, which is beneficial to produce robust patch embeddings and accurate sketch representations.

Furthermore, we enforce a clustering constraint on the embeddings of patches cropped from different sketches jointly with the network learning. The synonymous inter-sketch patches (e.g., A and C in Fig. \ref{fig:definition of icon synonym}) are encoded close together as clusters, and their embeddings are forced to move towards their assigned cluster centroids, which are learned from the entire training set. As a result, the embeddings become more compact and self-organized, and thus the computation of synonymous proximity is more accurate and robust against the variation in sketch drawings.

To realize the above idea, we propose synonymous proximity based graph to sequence (SP-gra2seq))\footnote{The codes are in: \url{https://github.com/CMACH508/SP-gra2seq}.}  to construct learnable patch linkings for graphic sketch representation. A sketch is cropped into patches which are embedded by a convolutional neural network (CNN) encoder. We construct the sketch graph based on the synonymous proximity, which is computed by cosine similarity between the patch embeddings. A GCN encoder is devised to aggregate the patch features for the final sketch representation. Moreover, we cluster the inter-sketch patch embeddings jointly with the network training, and push the embeddings towards their assigned cluster centroids to stabilize the estimation of synonymous proximity. To summarize, we make the following contributions: 
\begin{itemize}
	\item[1.] We propose SP-gra2seq to learn graphic sketch representations by linking sketch patches by synonymous proximity. The constructed graph edges are learnable to adapt to the variation of sketch drawings.
	\item[2.] SP-gra2seq enforces a regularization by clustering the embeddings of inter-sketch patches. The regularization improves the computation of the synonymous proximity.
	\item[3.] Experimental results show that SP-gra2seq significantly improves the state-of-the-art performance on controllable sketch synthesis \citep{zang2021controllable} and sketch healing \citep{su2020sketchhealer,qi2022generative}.
\end{itemize}

\section{Related work}

\subsection{Representing Sketches by Different Formats}

A sketch can be represented in multiple formats by highlighting and storing clues from different views. If a sketch is represented by a sequence of pen stroke actions, its temporal features have been captured by the recurrent neural network (RNN)-based \citep{ha2017neural} or transformer-based \citep{lin2020sketch,ribeiro2020sketchformer} encoders. If a sketch is formed as a raster image, the spatial relationships among pixels have been extracted by CNNs \citep{chen2017sketch,tian2021relationship}. Moreover, both sketch formats have been used simultaneously for efficient representations \citep{song2018learning,choi2019sketchhelper,xu2020learning,li2022multistage}.


Recently, graph neural networks (GNNs) \citep{scarselli2008graph} and GCNs were utilized for sketch recognition \citep{yang2020s,xu2021multigraph,li2021efficient}, sketch segmentation \citep{yang2021sketchgnn,qi2022one}, image retrieval \citep{zhang2020zero} and sketch synthesis \citep{su2020sketchhealer,qi2021sketchlattice,qi2022generative}. They focused on learning graphic sketch representations by highlighting the proximity among different parts of a sketch. A sketch was represented by the cropped patches \citep{su2020sketchhealer,qi2022generative} or the latticed coordinates on the canvas \citep{qi2021sketchlattice} and were regarded as the graph nodes. These nodes were linked by graph edges based on either the temporal proximity following the sketch drawing order \citep{su2020sketchhealer,qi2022generative} or the spatial proximity revealed by the Euclidean distances \citep{qi2021sketchlattice}. However, these edge constructions rely on the inherent sketch attributes, which are specific to sketch individuals. The huge variation of sketch drawings may reduce the accuracy for their representations.

Different from the existing methods, we link sketch patches by the learnable synonymous proximity to adapt to the sketch diversities for accurate and robust sketch representations.

\subsection{Constraining the Distribution of Sketch Codes}

Constructing a proper latent structure for modeling the sketch data manifold contributes to efficient sketch representations. When the sketch codes are single Gaussian distributed, e.g., in sketch-rnn \citep{ha2017neural}, sketches in different categories are chaotically encoded in a latent cluster, reducing the representing performance on multi-categorized datasets. To make the codes more freely encoded, sketch-pix2seq \citep{chen2017sketch} removed the Kullback-Leibler (KL) divergence term from the objective. RPCL-pix2seq \citep{zang2021controllable} self-organized Gaussian mixture model (GMM) distributed codes to constrain sketches with similar patterns in a compact Gaussian component of GMM. 


Inspired by the aforementioned articles, we cluster and regularize the latent embeddings of the sketch patches jointly with the network training. The clustered embeddings are more compact and self-organized, and thus it raises the accuracy of the computed synonymous proximity.

\section{Methodology}

\begin{figure*}[!t]
  	\centering
  	\includegraphics[width=0.9\linewidth]{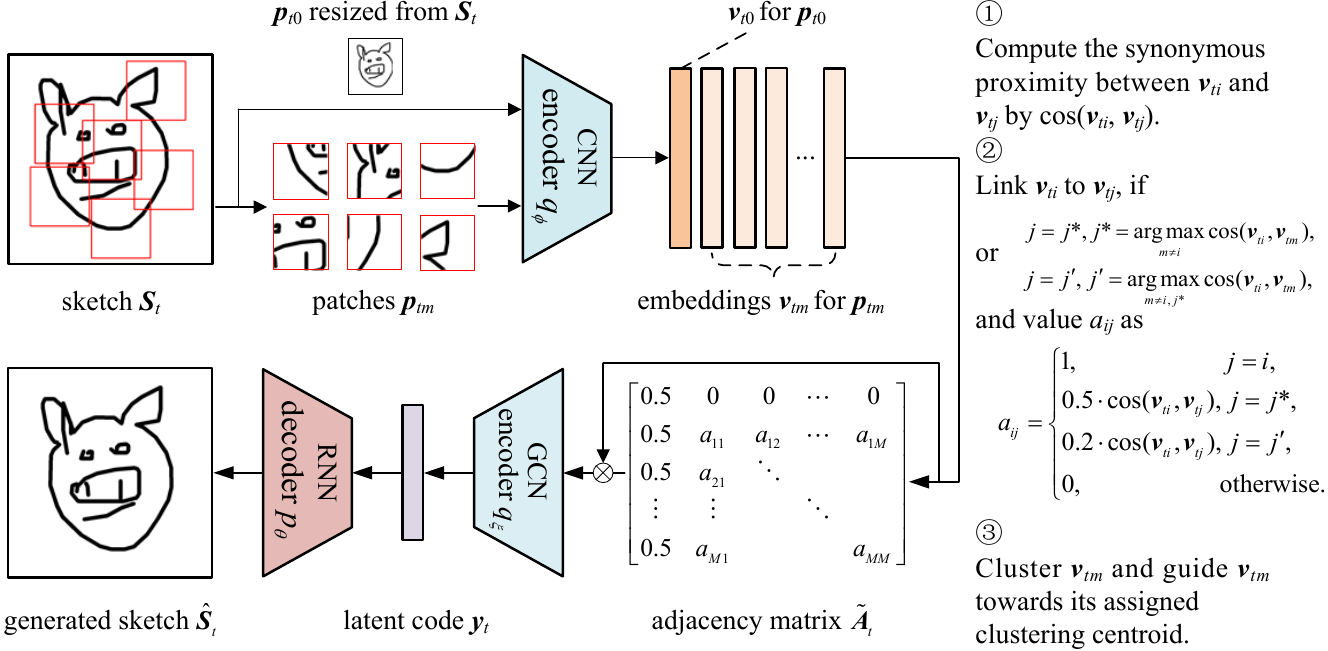}
  	\caption{An overview of SP-gra2seq. A sketch is cropped into patches which are embedded by a CNN encoder. The sketch graph is constructed based on the synonymous proximity computed by the patch embeddings. The features captured from patches are aggregated by a GCN encoder for the RNN decoder to reconstruct the input sketch.}
	\label{fig:overview}
\end{figure*}

Fig. \ref{fig:overview} offers an overview of SP-gra2seq. We firstly crop $M$ sketch patches $\{\bm{p}_{tm}|1\leq m\leq M\}$ from the $t$-th sketch $\bm{S}_t$ as the graph nodes and resize $\bm{S}_t$ to obtain $\bm{p}_{t0}$. They are fed into the CNN encoder $q_{\bm\phi}$ to compute their latent embeddings $\bm{v}_{tm}$ and $\bm{v}_{t0}$, respectively. We construct the graph edges based on the learnable synonymous proximity between the node embeddings. In practice, the adjacency matrix is obtained by using the top values of the cosine similarity between each node $\bm{v}_{ti}$ and its neighbors $\bm{v}_{tj}$ ($|1\leq i, j \leq M$). A GCN encoder $q_{\bm\xi}$ computes the final code $\bm{y}_t$ for the RNN decoder $p_{\bm{\theta}}$ to generate the reconstruction $\hat{\bm S}_t$. Furthermore, we cluster all patch embeddings $\{\bm{v}_{tm}|\text{for any }t, m\}$ and push $\bm{v}_{tm}$ towards its assigned clustering centroid.

\subsection{Linking Sketch Patches by Learnable Synonymous Proximity}

The \textbf{graph nodes} of a sketch are defined as the sketch patches following \citep{su2020sketchhealer,qi2022generative}. Firstly, we crop $M$ patches $\bm p_{tm}\in\mathbb{R}^{256\times 256}$ ($1\leq m\leq M$) from the sketch image $\bm S_t\in\mathbb{R}^{640\times 640}$. The selecting rule for the cropping positions is adopted from \citep{su2020sketchhealer,qi2022generative}, but we enlarge the patch size from $128\times 128$ as in \citep{su2020sketchhealer,qi2022generative} to $256\times 256$, ensuring that our patches collect enough drawing strokes for composing meaningful sketch parts, e.g., a cropped wheel in Fig. \ref{fig:definition of icon synonym}. 

Two patches with similar semantic contents or with analogous geometric shape of the drawing strokes are synonymous. For example, the patches A, B and C in Fig. \ref{fig:definition of icon synonym} are in close synonymous proximity of one another, as all of them collect a wheel from the vehicles. We employ a CNN encoder $q_{\bm\phi}(\bm v|\bm p)$ to capture the embeddings $\bm v_{ti}$ and $\bm v_{tj}\in\mathbb{R}^{512\times 1}$ ($1\leq i, j \leq M$) from the patches $\bm p_{ti}$ and $\bm p_{tj}$, respectively. $q_{\bm\phi}$ consists of seven convolutional layers (channel numbers as 8, 32, 64, 128, 256, 512 and 512) with the kernel size $2\times 2$ and the ReLU activation function, followed by max pooling and batch normalization. And we compute the cosine similarity to measure their \textbf{synonymous proximity}:
\begin{equation}
	\label{eq:cos}
	\cos(\bm v_{ti}, \bm v_{tj})=\frac{\bm v_{ti}^\text{T} \bm v_{tj}}{||\bm v_{ti}||_2\cdot||\bm v_{tj}||_2},
\end{equation}
where $||\cdot||_2$ denotes the L2 norm. A large value of $\cos(\bm v_{ti}, \bm v_{tj})$ indicates that the corresponding $\bm p_{ti}$ is in close synonymous proximity to $\bm p_{tj}$. As the patch embeddings are in a high dimensional space (e.g., 512 dimensions in this paper), the ratio of the Euclidean distances of the nearest and farthest neighbors to a given target is almost $1$ \citep{beyer1999nearest}. Hence, we use the cosine similarity as the metric.

Cropped from the same sketch $\bm S_t$, if $\bm p_{ti}$ and $\bm p_{tj}$ are in close synonymous proximity, we link them with a \textbf{graph edge}, e.g., linking A and B in Fig. \ref{fig:definition of icon synonym}. We store the computed synonymous proximity in an adjacency matrix $\bm A_t\in \mathbb{R}^{M\times M}$. The element $a_{ij}$ in $\bm A_t$ denotes the synonymous relationships between $\bm v_{ti}$ and $\bm v_{tj}$, which is
\begin{equation}
	\label{eq:element_a}
	a_{ij}=\left\{
			\begin{array}{ll}
				1, & j=i,\\
        			0.5\cdot\cos(\bm v_{ti}, \bm v_{tj}), & j=j^*,\\
				& j^*=\mathop{\arg\max}\limits_{m\neq i}\cos(\bm v_{ti}, \bm v_{tm}),\\
        			0.2\cdot\cos(\bm v_{ti}, \bm v_{tj}), & j=j',\\
				& j'=\mathop{\arg\max}\limits_{m\neq i, j^*}\cos(\bm v_{ti}, \bm v_{tm}),\\
        			0, & \text{otherwise.}
    			\end{array}\right.
\end{equation}

For a graph with $M$ nodes, the node $\bm v_{ti}$ is linked to $\bm v_{tj^*}$ and $\bm v_{tj'}$, which are with the top two values of cosine similarity among $\bm v_{tm}$ ($m\neq i$). Accordingly, $a_{ij}$ is proportional to the value of cosine similarity. As $\bm v_{tj^*}$ is more synonymously related to $\bm v_{ti}$ than $\bm v_{tj'}$, it is offered with a larger coefficient $0.5$ to transport more messages from $\bm v_{tj^*}$ to the target $\bm v_{ti}$. We only select the top two and weight them with 0.5 and 0.2 by their importance, making the adjacency matrix effective and sparse as in SketchHealer \citep{su2020sketchhealer} and SketchHealer 2.0 \citep{qi2022generative}. Moreover, each node $\bm v_{ti}$ is added with a self-connection, i.e., $a_{ii}=1$. The rest elements on the $i$th row of $\bm A_t$ are valued as $0$, indicating that these corresponding patches are less synonymous to $\bm p_{ti}$. The graph edges are constructed by the learnable synonymous proximity, according to the dynamic patch embeddings produced from the CNN encoder. It allows SP-gra2seq to automatically link the sketch patches for message passing during the graphic sketch representation learning.

\subsection{Graphic Sketch Representation}\label{sec:discuss_m_top}

\begin{figure*}[!t]
  	\centering
  	\includegraphics[width=0.9\linewidth]{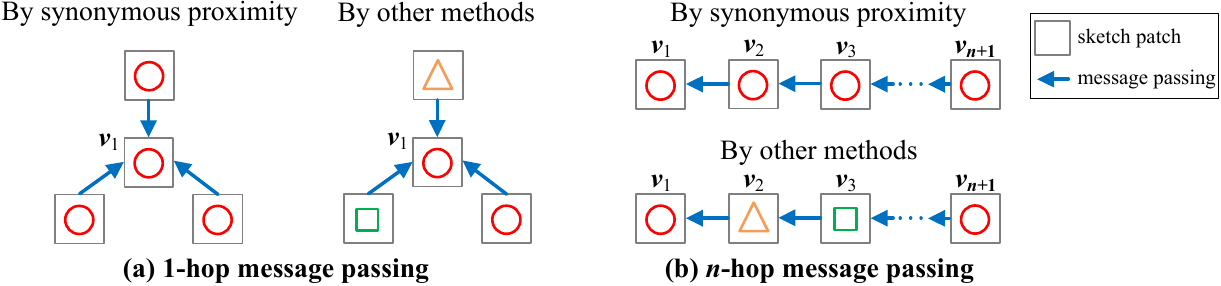}
  	\caption{Message aggregation by GCNs for learning the patch embedding $\bm v_1$. We use simple geometric shapes, e.g., circle, triangle, square, for convenience to roughly categorize the structural patterns in a patch. (a) $1$-hop message passing. When using synonymous proximity, a patch with a circle is linked to other ones with circles. The target $\bm v_1$ receives the messages from other circles to achieve a robust circle representation. But in other methods, $\bm v_1$ may receive the messages from a square or a triangle, which could be noises to interrupt the circle representation. (b) $n$-hop message passing by using stacked GCN layers. Actually, two synonymous nodes $\bm v_1$ and $\bm v_{n+1}$ may not be linked for not being at the top-3 neighbors of each other. But they could be connected in a multi-hop path. With synonymous proximity, multi-hop message passing would strengthen $\bm v_1$. But in other methods, it is more likely to connect the patch nodes of different patterns (e.g., the square) in a long multi-hop path.}
	\label{fig:aggregate}
\end{figure*}

With the constructed sketch graph, we integrate the information of all patch nodes via the synonymous proximity for the final sketch representation. We also resize the full sketch $\bm{S}_t$ to obtain $\bm{p}_{t0}$, which is with the same size as $\bm p_{tm}$. Its captured embedding $\bm v_{t0}$ from the CNN encoder $q_{\bm\phi}$ is regarded as a global observation of $\bm S_t$ to cooperate with the local details via the sketch patches $\bm{p}_{tm}$.

We build the GCN encoder $q_{\bm\xi}(\bm y|\bm V, \tilde{\bm A})$ to compute the final latent code $\bm y_t$ for sketch $\bm S_t$. $\bm V_t=\left[\bm v_{t0}, \bm v_{t1}, \cdots, \bm v_{tM}\right]^\text{T}$ concatenates all the outputs from the CNN encoder and $\tilde{\bm A}_t\in \mathbb{R}^{(M+1)\times (M+1)}$ is a matrix shown in Eq. (\ref{eq:adj_mat}),
\begin{small}
\begin{equation}
	\label{eq:adj_mat}
	\tilde{\bm A}_t=\left[
		\begin{array}{cc}
			0.5 & \bm 0^{\text{T}}\\
			0.5\cdot\bm 1 & \bm A_t
		\end{array}\right],
		\bm A_t=\left[
			\begin{array}{cccc}
				a_{11} & a_{12} & \cdots & a_{1M}\\
				a_{21} & a_{22} & \cdots & a_{2M}\\
				\vdots & \vdots & \ddots & \vdots\\
				a_{M1} & a_{M2} & \cdots & a_{MM}\\
			\end{array}\right],
\end{equation}
\end{small}
where $\bm 0$ and $\bm 1$ are two $M\times 1$ vectors with all elements valued as $0$ and $1$, respectively. And the element $a_{ij}$ in $\bm A_t$ is computed by Eq. (\ref{eq:element_a}).



We construct a GCN layer with the weight $\bm W$ as
\begin{equation}
	\label{eq:norm_adj}
	\bm F_t=\text{ReLU}\left(\tilde{\bm D}_t^{-\frac{1}{2}}\tilde{\bm A}_t\tilde{\bm D}_t^{-\frac{1}{2}}\bm V_t\bm W\right),
\end{equation}
where $\tilde{\bm D}_t$ represents the degree matrix of $\tilde{\bm A}_t$. The embedding $\bm v_{t0}$ extracted from the full sketch is weighted by the split first column of $\tilde{\bm A}_t$ with the weights $0.5$ in Eq. (\ref{eq:adj_mat}), cooperating with $\{\bm v_{tm}\}_{m=1}^M$ from the sketch patches to yield a comprehensive feature $\bm F_t$. $\bm F_t$ is further sent into a pair of fully connected layers to obtain two vectors $\bm\mu_t$ and $\bm\sigma_t$, which are for computing the final code $\bm y_t$ by $\bm y_t=\bm\mu_t+\bm\sigma_t\odot\bm\epsilon$. The operation $\odot$ denotes the Hadamard product and $\bm\epsilon$ is randomly sampled from a standard Gaussian distribution $\mathcal{G}(\bm\epsilon|\bm 0, \bm I)$. 

We illustrate the relative advantage of our method which is driven by synonymous proximity in Fig. \ref{fig:aggregate}, where for convenience we use simple geometric shapes, e.g., circle, triangle, square, to roughly categorize the structural patterns in a patch. For example, in Fig. \ref{fig:aggregate}(a), the target node $\bm v_1$ representing a circle receives the messages from its neighbors. If the graph edges are constructed by our synonymous proximity, aggregating the messages from the neighboring patches of circle plays a role of denoising, which is beneficial to produce robust and accurate embedding of $\bm v_1$. However, if the edges are built by other methods, e.g., the temporal proximity, the passing messages may bring noises (e.g., messages from the square and the triangle) to interrupt the circle representation.

The message passing can form multi-hop paths after going through the stacked GCN layers by Eq. (\ref{eq:norm_adj}). Although two patch nodes, like $\bm v_1$ and $\bm v_{n+1}$ in Fig. \ref{fig:aggregate}(b), which are actually very similar to each other, may not have an edge between them for not being at the top-3 neighbors of each other according to Eq. (\ref{eq:element_a}), they are still likely to be connected in a multi-hop path. With synonymous proximity, the multi-hop message passing will strengthen the patch embeddings $\bm v_1$ and lead to accurate sketch representations. When using temporal proximity or spatial proximity, it is even more likely to connect the patch nodes of different patterns (e.g., the square and the triangle) in a long multi-hop path, leading to inconsistent, corrupted information propagation.


\subsection{Clustering Sketch Patches from Different Sketches}

Patches cropped from different sketches can be in close synonymous proximity of one another as well, e.g., the patches A and C in Fig. \ref{fig:definition of icon synonym}. Though these inter-sketch patches cannot be linked by edges, we enforce a clustering constraint on their embeddings jointly with the network training.

At the ($\tau$)-th training iteration, the network is fed with a mini-batch of $N$ sketches $\{\bm S_t\}_{t=1}^N$. Thus, we capture all $N\times M$ embeddings $\bm v_{tm}$ and cluster them in the latent space.
\begin{small}
\begin{align}
	&q^{(\tau)}(k|\bm v_{tm})=\left\{
					\begin{array}{ll}
						1, &k=\mathop{\arg\max}\limits_{k^*}\cos(\bm v_{tm}, \bm c^{(\tau-1)}_{k^*}),\\
						0, & \text{otherwise.}
					\end{array}\right.\nonumber\\
	&\bm c^{(\tau)}_k=\eta\cdot\frac{\sum_{t=1}^N\sum_{m=1}^Mq^{(\tau)}(k|\bm v_{tm})\cdot\bm v_{tm}}{\sum_{t=1}^N\sum_{m=1}^Mq^{(\tau)}(k|\bm v_{tm})}+(1-\eta)\cdot\bm c^{(\tau-1)}_k,\label{eq:em_step}
\end{align}
\end{small}
where $\bm c_k\in\mathbb{R}^{512\times 1}$ ($1\leq k\leq K$) and $\eta$ denote the cluster centroids and the learning rate for updating $\bm c_k$, respectively. The node $\bm v_{tm}$ is assigned to the $k$th cluster, if $\bm v_{tm}$ and $\bm c_k$ are with the largest cosine similarity among all $K$ centroids. 

The nodes $\{\bm v_{tm}\}$ with analogous patterns are assigned to the same cluster, wherever $\bm v_{tm}$ is from. Hence, the cluster centroid $\bm c_k$ generalizes the synonymous features from the entire training set. We guide $\bm v_{tm}$ to move towards its assigned cluster centroid to self-organize the compact embeddings, and thus the computation of the synonymous proximity is more robust against the variation of sketch drawings.

\subsection{Training an SP-gra2seq via Sketch Reconstruction}

We send the final code $\bm y_t$ for sketch $\bm S_t$ into the RNN decoder $p_{\bm\theta}(\bm S|\bm y)$, whose network architecture is adopted from sketch-rnn \citep{ha2017neural}, to reconstruct the sketch $\hat{\bm S}_t$ in a sequence format. Our objective is to maximize
\begin{align}
	\label{eq:loss}
	\mathcal{L}(\bm\theta, \bm\xi, &\bm\phi|\bm S)=\sum_{t=1}^N\left[\vphantom{\sum_i^N}\text{E}_{q_{\bm\phi, \bm\xi}(\bm y_t|\bm S_t)}\left[\log p_{\bm\theta}(\bm S_t|\bm y_t)\right]\right.\nonumber\\
	&\left.-\lambda\sum_{m=1}^M\sum_{k=1}^Kq(k|\bm v_{tm})\cdot||\bm v_{tm}-\text{sg}(\bm c_k)||_2\right],
\end{align}
where $\text{sg}(\cdot)$ stands for the stop-gradient operator. The first log-likelihood term in Eq. (\ref{eq:loss}) aims to reconstruct the input by the sequence-formed sketch generation, and we calculate this term following \cite{ha2017neural}. The second term weighted by $\lambda$ is a regularization by pushing $\bm v_{tm}$ towards its assigned cluster centroid $\bm c_k$. We remove the KL divergence term $\text{KL}\left[q_{\bm\phi, \bm\xi}(\bm y|\bm S_t)||p(\bm y_t)\right]$ from the objective \citep{chen2017sketch,su2020sketchhealer,qi2021sketchlattice} to encourage the latent code $\bm y_t$ to be more freely encoded.

\section{Experiments}

We choose controllable sketch synthesis \citep{zang2021controllable} and sketch healing \citep{su2020sketchhealer,qi2022generative} to testify whether our method learns accurate and robust graphic sketch representations. 

\subsection{Preparation}

\begin{figure*}[ht]
 	\centering
 	\includegraphics[width=\linewidth]{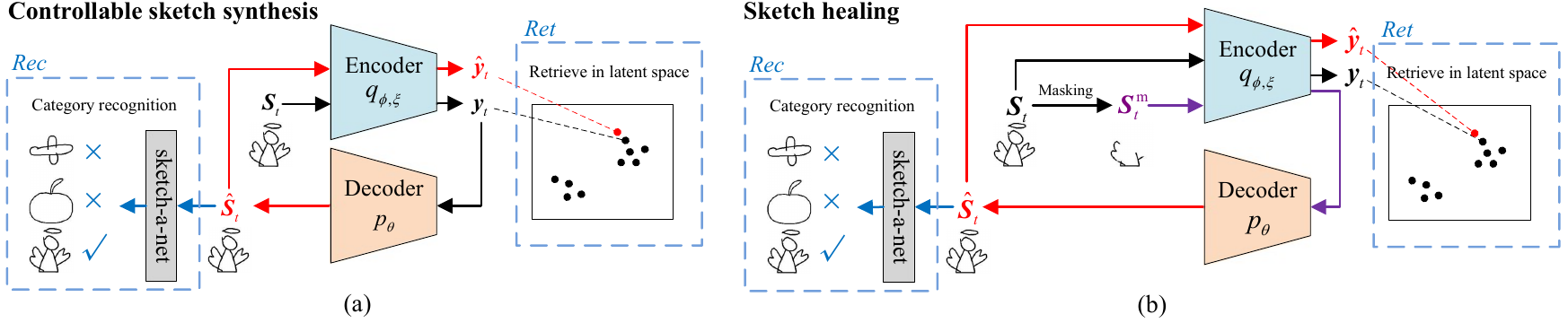}
 	\caption{Calculating $Rec$ and $Ret$ for (a) controllable sketch synthesis and (b) sketch healing.}
	\label{fig:metrics}
\end{figure*}

\noindent\textbf{Datasets}. Three datasets from \emph{QuickDraw} \citep{ha2017neural} are selected for experimental comparison. Dataset 1 (DS1) and dataset 2 (DS2) are adopted from \cite{zang2021controllable} and \cite{qi2021sketchlattice}, respectively. DS1 (bee, bus, flower, giraffe and pig) evaluates the representation learning on the large variations of the sketches within the same category. DS2 (airplane, angel, apple, butterfly, bus, cake, fish, spider, the Great Wall and umbrella) evaluates whether the methods are sensitive to categorical patterns. We also introduce the most challenging dataset 3 (DS3), which is an enhanced version of DS1 with three newly introduced categories (car, cat and horse). DS3 evaluates whether the methods can still recognize the sketch categories from the shared non-categorical patterns between categories (e.g., wheels can be found in cars or buses). Each category contains $70K$ training, $2.5K$ valid and $2.5K$ test samples ($1K=1000$).

\begin{table*}[!t]
	\centering
	\begin{tabular}{lcrrrcrrrcrrr}
		\hline
    		\multirow{3}*{Models} & \multicolumn{4}{c}{DS1} & \multicolumn{4}{c}{DS2} & \multicolumn{4}{c}{DS3}\\ \cline{2-13}
		~ & \multirow{2}*{$Rec$} & \multicolumn{3}{c}{$Ret$ (top-)} & \multirow{2}*{$Rec$} & \multicolumn{3}{c}{$Ret$ (top-)} & \multirow{2}*{$Rec$} & \multicolumn{3}{c}{$Ret$ (top-)}\\ \cline{3-5} \cline{7-9} \cline{11-13}
		~ & ~ & 1 & 10 & 50 & ~ & 1 & 10 & 50 & ~ & 1 & 10 & 50\\
    		\hline
		sketch-rnn & 50.33 & 0.38 & 2.84 & 9.33 & 46.28 & 10.93 & 23.73 & 48.38 & 57.64 & 3.72 & 13.42 & 26.14\\
		sketch-pix2seq & 83.99 & 13.45 & 30.12 & 49.99 & 85.46 & 50.94 & 71.38 & 80.15 & 79.13 & 22.92 & 47.55 & 58.19\\
		Song et al. & 91.77 & 16.41 & 36.43 & 52.22 & 86.98 & 58.84 & 76.84 & 80.06 & 83.28 & 25.47 & 43.39 & 56.16\\
		RPCL-pix2seq & 93.18 & 17.86 & 38.87 & 55.30 & 88.73 & 53.19 & 71.60 & 87.91 & 81.80 & 28.80 & 59.05 & 77.52\\
		SketchHealer & 91.04 & 58.80 & 82.15 & 91.33 & 94.04 & 87.54 & 96.19 & 98.26 & 87.03 & 68.52 & 82.37 & 86.57\\
		SketchHealer 2.0 & 93.13 & 57.19 & 84.54 & 90.26 & 90.94 & 87.37 & 94.59 & 97.60 & 87.37 & 50.67 & 76.11 & 82.42\\
		SketchLattice & 75.91 & 6.55 & 14.01 & 26.72 & 71.80 & 6.91 & 14.76 & 28.82 & 62.21 & 5.90 & 10.36 & 19.39\\
		SketchLattice$^+$ & 95.18 & 72.74 & 91.60 & 97.14 & 94.30 & 90.56 & 97.78 & \pmb{99.27} & 89.49 & 87.27 & 96.82 & 98.98\\
		SP-gra2seq & \pmb{95.91} & \pmb{94.88} & \pmb{99.11} & \pmb{99.72} & \pmb{94.85} &\pmb{90.83} & \pmb{98.29} & 99.08 & \pmb{89.83} & \pmb{94.05} & \pmb{98.72} & \pmb{99.57}\\
  		\hline
	\end{tabular}
	\caption{Controllable sketch synthesis performance (\%) on three datasets.}
	\label{tab:cs_performance}
\end{table*}

\noindent\textbf{Baselines}. We compare our SP-gra2seq with eight baseline models. Sketch-rnn \citep{ha2017neural} learns the sketch representations from sketch sequences. Sketch-pix2seq \citep{chen2017sketch} and RPCL-pix2seq \citep{zang2021controllable} both use sketch images as input, and constrain the sketch codes with a proper distribution to learn better representations. Song et al. \citep{song2018learning} is fed with sketch sequences and images as pairs simultaneously. SketchHealer \citep{su2020sketchhealer}, SketchLattice \citep{qi2021sketchlattice} and SketchHealer 2.0 \citep{qi2022generative} learn graphic sketch representations by linking graph nodes based on the temporal or spatial proximity. Especially, the source code of SketchHealer 2.0 is not released yet, and the article does not provide the details to develop a differentiable sketch rasterisation module for calculating the perceptual loss term. We employ a bi-directional long-short term memory (Bi-LSTM) to calculate the perceptual loss term over the sketch sequences. Besides, we use the Gumbel-softmax \citep{jang2016categorical} to make the sampling process differentiable. Moreover, we also produce an extended SketchLattice named SketchLattice$^+$, whose node embeddings are captured from sketch patches as in SketchHealer, instead of the coordinates on lattice. A lattice of $8\times 8$ is set on the canvas to determine the cropping positions, and the Euclidean distances between them are utilized to construct graph edges as in SketchLattice.

When training an SP-gra2seq, the patch number $M$, the mini-batch size $N$, the learning rate $\eta$ for updating clustering centroids and the weight $\lambda$ in the objective are fixed at $20$, $256$, $0.05$ and $0.25$, respectively. The numbers of cluster centroids $K$ are $30$, $50$ and $50$ for three datasets, respectively. We employ Adam optimizer for the network learning with the parameters $\beta_1=0.9$, $\beta_2=0.999$ and $\epsilon=10^{-8}$. And the learning rate starts from $10^{-3}$ with a decay rate of $0.95$ for each training epoch.

\subsection{Controllable Sketch Synthesis}\label{sec:controllable}

Controllable sketch synthesis requires the method to generate sketches $\hat{\bm S}_t$ with exact patterns as the input $\bm S_t$. We quantitatively evaluate the performance by metrics $Rec$ and $Ret$ proposed by \cite{zang2021controllable}. The calculations of $Rec$ and $Ret$ are in Fig. \ref{fig:metrics}(a). $Rec$ indicates whether the generated sketch $\hat{\bm S}_t$ and its corresponding input $\bm S_t$ are in the same category. We respectively pre-train three sketch-a-net \citep{yu2017sketch} as classifiers to calculate $Rec$ for three datasets. $Ret$ represents whether $\hat{\bm S_t}$ is well controlled to preserve both the categorical and the detailed non-categorical patterns from $\bm S_t$. More specifically, when feeding the network with $\bm S_t$, we obtain its latent code $\bm y_t$ and the generated sketch $\hat{\bm S}_t$. By sending $\hat{\bm S}_t$ back to the same encoder, we get its corresponding code $\hat{\bm y}_t$. We retrieve the original $\bm y_t$ from $\bm Y=\{\bm y_t(\bm S_t)|\bm S_t\in\text{test set}\}$ with $\hat{\bm y}_t$, and $Ret$ is the successful retrieving rate. Both $Rec$ and $Ret$ are calculated from the entire test set.

Table \ref{tab:cs_performance} reports the controllable sketch synthesis performance. Fed with the paired sketch sequences and images, the sketch representations learned by Song et al. \citep{song2018learning} collect both the temporal drawing orders and spatial dependencies among pixels. It achieves higher performance than sketch-rnn \citep{ha2017neural} and sketch-pix2seq \citep{chen2017sketch}, which only use either the sketch sequences or images. RPCL-pix2seq \citep{zang2021controllable} self-organizes the sketch codes in clusters by their patterns, which is also effective to raise the performance.

SketchHealer \citep{su2020sketchhealer}, SketchHealer 2.0 \citep{qi2022generative} and SketchLattice$^+$ achieve significant improvement on $Ret$ by learning graphic sketch representations. The temporal or spatial proximity for constructing sketch graphs improves grasping the sketch details. Thus, the generated sketches resemble the corresponding input, leading to high $Ret$ performance. Especially, SketchHealer 2.0 uses a perceptual loss term to guide the network to preserve more global semantics of sketches, instead of exactly reconstructing the local input details. As these two objectives are trade-off \citep{qi2022generative}, SketchHealer outperforms SketchHealer 2.0 a little on controllable synthesis. Besides, the original SketchLattice \citep{qi2021sketchlattice} is a light weight model by learning the representations from the coordinates only. SketchLattice fails to restore the sketch details in the generations, resulting in poor performance on $Ret$.

The graph edges constructed by SP-gra2seq are equipped with synonymous awareness, which are more adaptive to the divergent sketch drawings, especially when sketches are rich in non-categorical patterns, e.g., in DS3. The passing messages through the edges are more reliable to produce accurate patch embeddings. Hence, SP-gra2seq achieves the best performance on controllable synthesis.

\begin{figure*}[!t]
  	\centering
  	\includegraphics[width=0.95\linewidth]{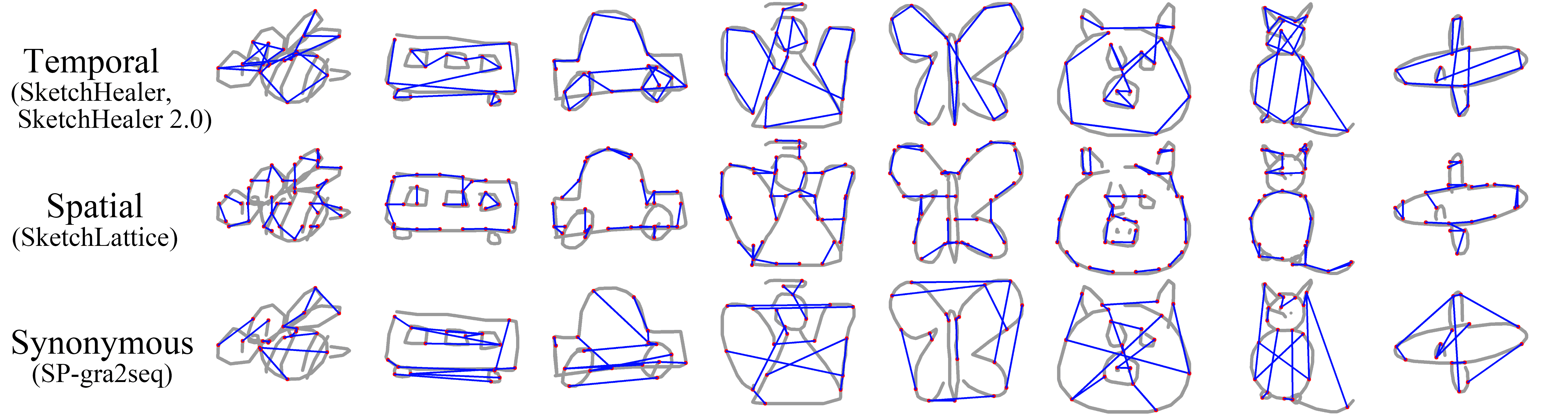}
  	\caption{The constructed edge links by temporal, spatial and the proposed synonymous proximity. The red dots denote the patching centers (or the coordinates on lattice) as graph nodes. In this figure, each node is linked to its ``nearest'' neighboring node with a blue line by the calculated proximity, representing the graph edge.}
	\label{fig:graph_edge}
\end{figure*}

\begin{table*}[!t]
	\centering
	\small
	\begin{tabular}{clcrrrcrrrcrrr}
		\hline
    		\multirow{3}*{Mask} & \multirow{3}*{Models} & \multicolumn{4}{c}{DS1} & \multicolumn{4}{c}{DS2} & \multicolumn{4}{c}{DS3}\\ \cline{3-14}
		~ & ~ & \multirow{2}*{$Rec$} & \multicolumn{3}{c}{$Ret$ (top-)} & \multirow{2}*{$Rec$} & \multicolumn{3}{c}{$Ret$ (top-)} & \multirow{2}*{$Rec$} & \multicolumn{3}{c}{$Ret$ (top-)}\\ \cline{4-6} \cline{8-10} \cline{12-14}
		~ & ~ & ~ & 1 & 10 & 50 & ~ & 1 & 10 & 50 & ~ & 1 & 10 & 50\\
    		\hline
		\multirow{5}*{10\%} & SketchHealer & 70.38 & 14.25 & 27.91 & 45.51 & 70.56 & 35.22 & 55.86 & 66.24 & 60.91 & 15.10 & 37.89 & 53.10\\
		~ & SketchHealer 2.0 & 90.18 & 19.45 & 41.51 & 59.94 & 87.07 & 39.40 & 64.01 & 81.20 & 76.45 & 15.55 & 39.99 & 61.67\\
		~ & SketchLattice & 54.18 & 1.09 & 5.48 & 14.00 & 52.97 & 1.34 & 6.86 & 17.41 & 44.14 & 0.71 & 4.19 & 11.36\\
		~ & SketchLattice$^+$ & 89.98 & 19.86 & 43.59 & 63.70 & 90.52 & 46.98 & 70.21 & 83.12 & 78.28 & 22.56 & 44.89 & 62.11\\
		~ & SP-gra2seq & \pmb{92.90} & \pmb{41.24} & \pmb{65.74} & \pmb{80.16} & \pmb{91.24} & \pmb{50.42} & \pmb{73.35} & \pmb{85.18} & \pmb{83.38} & \pmb{40.20} & \pmb{63.40} & \pmb{77.52}\\
		\hdashline[2pt/2pt]
		\multirow{5}*{30\%} & SketchHealer & 59.00 & 0.23 & 3.48 & 10.76 & 61.26 & 7.98 & 19.04 & 35.03 & 48.90 & 0.43 & 7.36 & 15.79\\
		~ & SketchHealer 2.0 & 79.05 & 4.66 & 14.54 & 28.44 & 75.08 & 10.05 & 24.51 & 39.57 & 60.74 & 3.75 & 11.66 & 22.36\\
		~ & SketchLattice & 32.03 & 0.08 & 2.05 & 5.96 & 31.73 & 0.88 & 3.64 & 9.25 & 23.06 & 0.41 & 2.41 & 5.96\\
		~ & SketchLattice$^+$ & 70.91 & 1.02 & 5.26 & 14.28 & 81.76 & 10.43 & 26.44 & 42.90 & 67.31 & 2.75 & 11.13 & 23.68\\
		~ & SP-gra2seq & \pmb{84.85} & \pmb{12.87} & \pmb{29.39} & \pmb{45.58} & \pmb{82.85} & \pmb{12.19} & \pmb{29.37} & \pmb{46.53} & \pmb{71.07} & \pmb{5.65} & \pmb{17.40} & \pmb{32.90}\\
  		\hline
	\end{tabular}
	\caption{Sketch healing performance (\%) on three datasets. ``Mask'' denotes the probability for the sketch patches to be masked.}
	\label{tab:sh_performance}
\end{table*}

Moreover, Fig. \ref{fig:graph_edge} presents a comparison between different principles for constructing the graph edges. When using the synonymous proximity, the blue lines regarded as the graph edges can cross the entire canvas to link the most synonymous patches, e.g., the connected bus wheels and windows in the $2$nd column and the connected butterfly wings in the $5$th column. The passing messages are more freely activated for efficient and accurate sketch representations. But these sketch parts fail to be linked by temporal or spatial proximity, as they are temporally separated in drawing order or with huge distances on the canvas.

\subsection{Sketch Healing}\label{sec:healing}

Sketch healing requires the method to recreate a full sketch $\hat{\bm S}_t$ from a corrupted sketch $\bm S^\text{m}_t$ with masks. If $\hat{\bm S}_t$ resembles the original $\bm S_t$ ($\bm S^\text{m}_t$ is corrupted from $\bm S_t$), it is regarded as a successful sketch healing. We use the similar metrics $Rec$ and $Ret$ for evaluating sketch healing, but adjust the calculating rule of $Ret$ to fit this task, as shown in Fig. \ref{fig:metrics}(b). It is worth mentioning that our retrieval (calculating $Ret$) is totally different from the ones in the articles of SketchHealer, SketchHealer 2.0 and SketchLattice. Firstly, we expect to retrieve, more exactly, identify the unique sketch $\bm S_t$ with $\hat{\bm S}_t$. We requires the sketch representations to preserve not only the categories, but also the non-categorical details for sketch identification. But these articles only retrieve sketches of the same category from the gallery. Secondly, our $Ret$ is calculated on the entire test set. Each sketch from the test set ($2500$ sketches per category) is selected in the query by turns with the entire test set as the gallery, while the retrieval in SketchHealer or SketchHealer 2.0 is only computed on $500$ sketches per category.


\begin{figure*}[ht]
 	\centering
 	\includegraphics[width=0.9\linewidth]{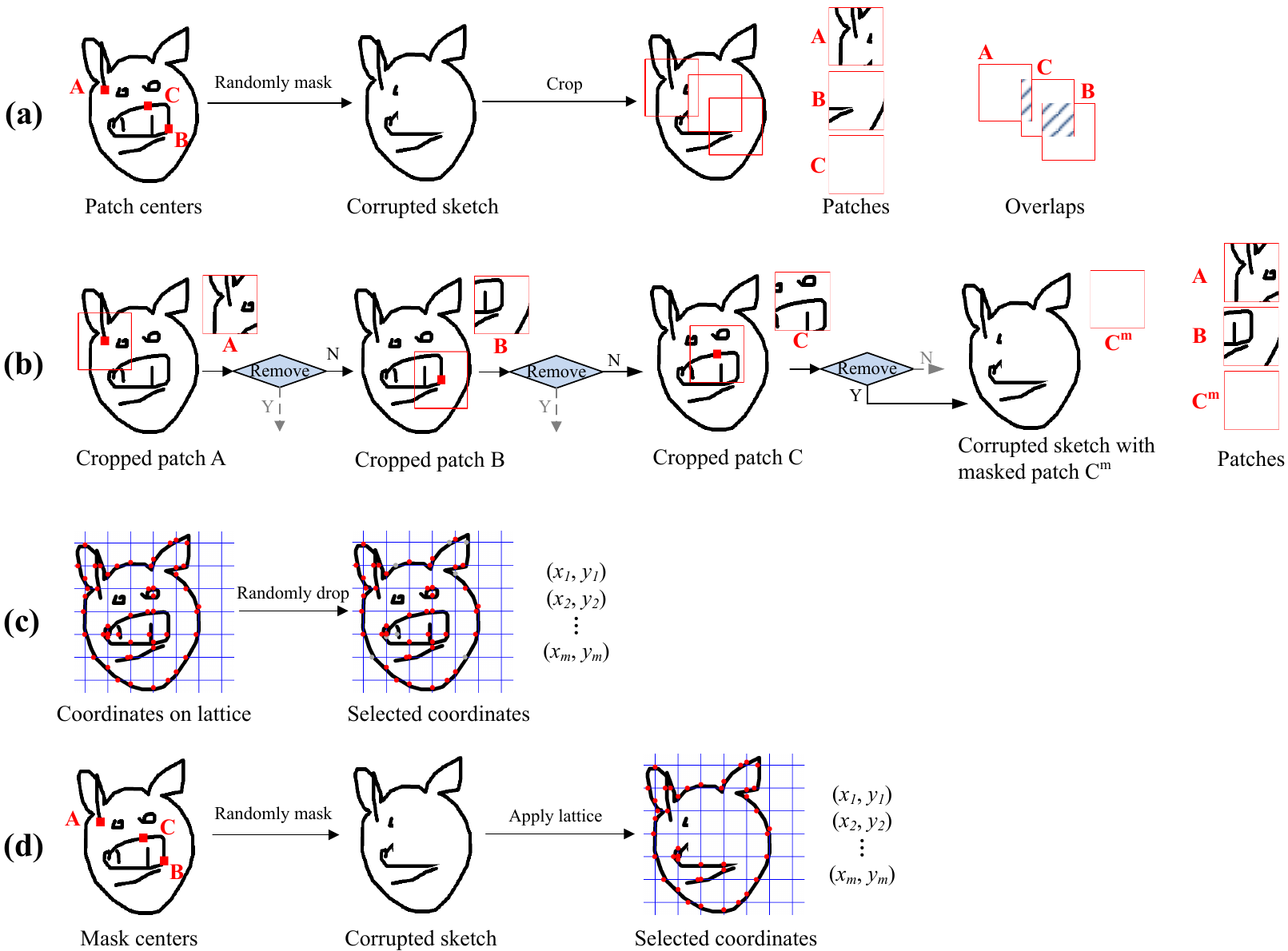}
 	\caption{Creating corrupted sketches by (a) our approach utilized in the paper, (b) the approach utilized in SketchHealer \citep{su2020sketchhealer} from the officially released codes in \url{https://github.com/sgybupt/SketchHealer}, (c) the approach utilized in SketchLattice \citep{qi2021sketchlattice} and (d) our approach adjusted for SketchLattice.}
	\label{fig:create_masks}
\end{figure*}

\noindent\textbf{Approaches for creating masks}. Our approach for masking sketches is different from SketchHealer, SketchHealer 2.0 and SketchLattice. Fig. \ref{fig:create_masks} presents four different approaches for creating corrupted sketches for sketch healing. In Fig. \ref{fig:create_masks}(a), we separate the patch cropping from the canvas masking in the pipeline. After positioning all patch centers on the canvas, we randomly select the patch centers (e.g., the patch C in the sub-figure) by a probability ($10\%$ or $30\%$) and remove their corresponding patches, i.e., masking. After all the selected patches are removed, we finally crop  patches at the same patch centers from the corrupted canvas. The graph edges linked to the masked patches are cut off as well. Especially, the patches A and C, the patches B and C are with overlapped regions, respectively, but no additional information below the masked patch C are leaked out to neither the patch A nor B.

For the masking approach of SketchHealer shown in Fig. \ref{fig:create_masks}(b), cropping and masking patches are applied by turns with the sketch drawing order. When two patches B and C share an overlapped region, B is cropped in front of C without being masked, but C is masked. The pixels located in the overlap leak out to the patch B, making the corrupted sketches much easier to be represented.

In Fig. \ref{fig:create_masks}(c), SketchLattice firstly creates a lattice on the sketch canvas and obtains all the coordinates, which are the overlapping points between strokes and lattice. Then it applies a point-level masking by randomly dropping a fraction of lattice points (the gray points are dropped) to determine the finally selected coordinates for learning graphic representation. The masked region (masked points exactly) is much smaller than ours by patch-level masking.

We also adjust our masking approach for SketchLattice, shown in Fig. \ref{fig:create_masks}(d), ensuring that the corrupted sketches fed to SketchLattice share the same corrupting level with other models. The sketch masking and coordinate selecting are separately applied by two steps. More specifically, the lattice is created after the sketch masking, and more coordinates may be dropped comparing with Fig. \ref{fig:create_masks}(c).

Table \ref{tab:sh_performance} reports the sketch healing performance. Light weight SketchLattice is heavily sensitive to the coordinate selection due to the limited information input. If the generated sketches contain improper strokes (e.g., the redundant circular strokes), the overlapping points between the improper strokes and the lattice are counted as corruptions, resulting in inaccurate sketch representations. 

SketchHealer 2.0 focuses on preserving the global semantics from the corrupted sketches by applying the perceptual loss in the objective, instead of exactly reconstructing the local details. Accordingly, when the sketches are corrupted with a large masking probability, e.g., $30\%$, SketchHealer 2.0 can still recognize the categorical and the semantic contents from scratch and generates resembled sketches.

For a partially masked patch, SP-gra2seq seeks its synonymous neighbors and constructs their edge linkings. The passing messages may contain the missing information under the masks, which benefits the patch embedding learning to be more comprehensive and accurate. Besides, as the patch embeddings are self-organized in compact clusters, a corrupted patch could still be encoded near the cluster centroid to make the representation robust. Thus, SP-gra2seq achieves the best sketch healing performance.

\begin{figure*}[!t]
  	\centering
  	\includegraphics[width=0.7\linewidth]{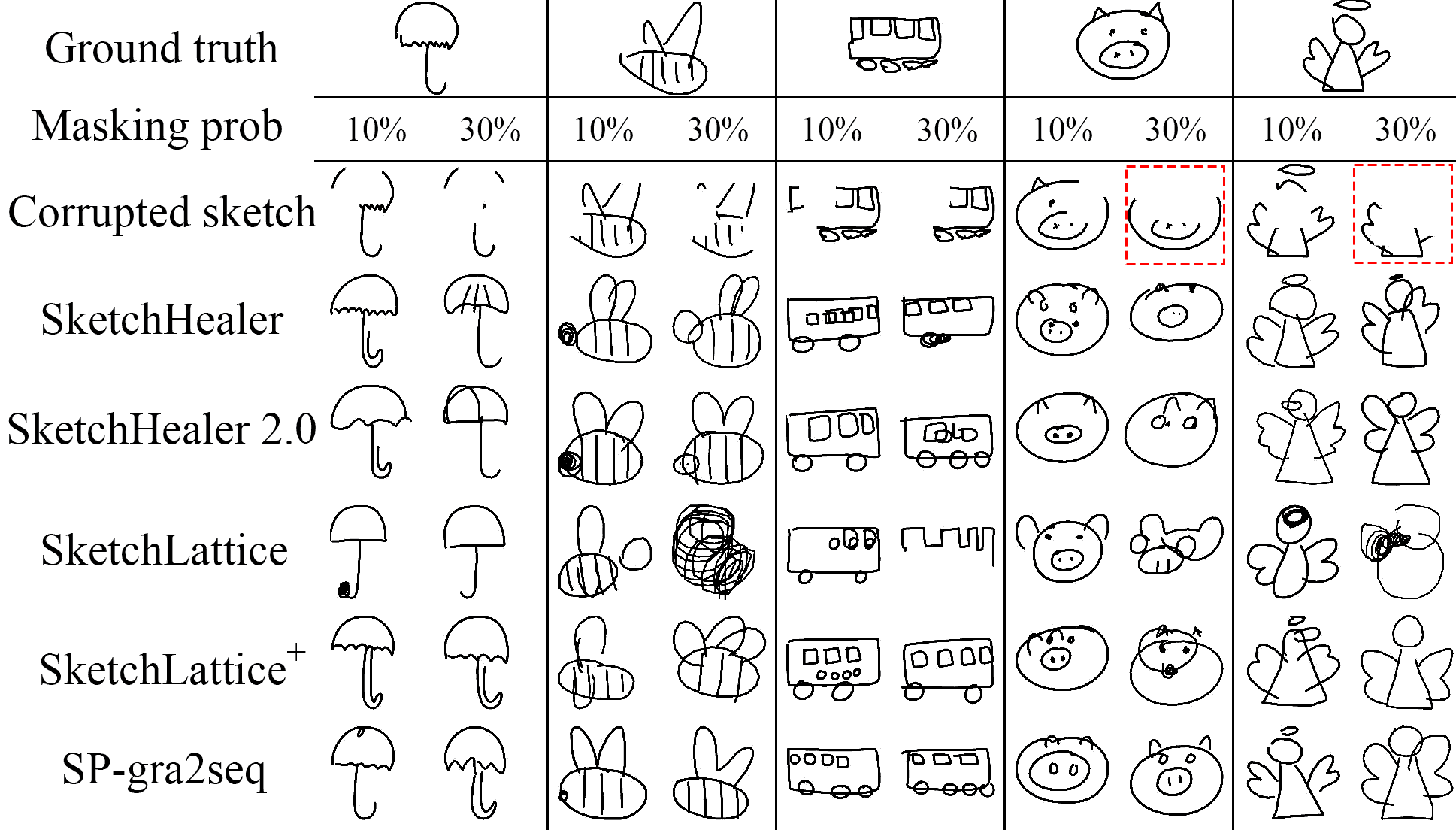}
  	\caption{Exemplary sketch healing results. The red dash boxes denote the sketches with masked key characteristics.}
	\label{fig:sketch_healing}
\end{figure*}

\begin{table*}[!t]
	\centering
	\small
	\begin{tabular}{cccrrrcrrrcrrr}
		\hline
    		\multirow{3}*{Mask} & \multirow{3}*{\makecell{Random\\linkings}} & \multicolumn{4}{c}{DS1} & \multicolumn{4}{c}{DS2} & \multicolumn{4}{c}{DS3}\\ \cline{3-14}
		~ & ~ & \multirow{2}*{$Rec$} & \multicolumn{3}{c}{$Ret$ (top-)} & \multirow{2}*{$Rec$} & \multicolumn{3}{c}{$Ret$ (top-)} & \multirow{2}*{$Rec$} & \multicolumn{3}{c}{$Ret$ (top-)}\\ \cline{4-6} \cline{8-10} \cline{12-14}
		~ & ~ & ~ & 1 & 10 & 50 & ~ & 1 & 10 & 50 & ~ & 1 & 10 & 50\\
    		\hline
		\multirow{2}*{0\%} & $\checkmark$ & 95.50 & 66.79 & 87.81 & 95.01 & 90.23 & 31.56 & 61.80 & 80.22 & 86.90 & 56.88 & 80.57 & 90.64\\
		~ & $\times$ & \pmb{95.91} & \pmb{94.88} & \pmb{99.11} & \pmb{99.72} & \pmb{94.85} &\pmb{90.83} & \pmb{98.29} & \pmb{99.08} & \pmb{89.83} & \pmb{94.05} & \pmb{98.72} & \pmb{99.57}\\
		\hdashline[2pt/2pt]
		\multirow{2}*{10\%} & $\checkmark$ & 91.57 & 38.91 & 64.67 & 79.52 & 87.35 & 18.53 & 42.98 & 63.13 & 82.63 & 32.22 & 56.32 & 72.01\\
		~ & $\times$ & \pmb{92.90} & \pmb{41.24} & \pmb{65.74} & \pmb{80.16} & \pmb{91.24} & \pmb{50.42} & \pmb{73.35} & \pmb{85.18} & \pmb{83.38} & \pmb{40.20} & \pmb{63.40} & \pmb{77.52}\\
		\hdashline[2pt/2pt]
		\multirow{2}*{30\%} & $\checkmark$ & 84.67 & 11.34 & 27.75 & 44.38 & 76.93 & 6.42 & 19.04 & 35.05 & 69.65 & \pmb{8.92} & \pmb{22.33} & \pmb{36.92}\\
		~ & $\times$ & \pmb{84.85} & \pmb{12.87} & \pmb{29.39} & \pmb{45.58} & \pmb{82.85} & \pmb{12.19} & \pmb{29.37} & \pmb{46.53} & \pmb{71.07} & 5.65 & 17.40 & 32.90\\
  		\hline
	\end{tabular}
	\caption{Controllable sketch synthesis (Mask~=~0\%) and sketch healing performance of SP-gra2seq by linking the sketch patches randomly or by the introduced synonymous proximity. ``Random linkings'' denotes the sketch patches are randomly linked.}
	\label{tab:ab_study_random}
\end{table*}

\begin{table*}[!t]
	\centering
	\small
	\begin{tabular}{cccrrrcrrrcrrr}
		\hline
    		\multirow{3}*{Mask} & \multirow{3}*{\makecell{Clustering\\constraint}} & \multicolumn{4}{c}{DS1} & \multicolumn{4}{c}{DS2} & \multicolumn{4}{c}{DS3}\\ \cline{3-14}
		~ & ~ & \multirow{2}*{$Rec$} & \multicolumn{3}{c}{$Ret$ (top-)} & \multirow{2}*{$Rec$} & \multicolumn{3}{c}{$Ret$ (top-)} & \multirow{2}*{$Rec$} & \multicolumn{3}{c}{$Ret$ (top-)}\\ \cline{4-6} \cline{8-10} \cline{12-14}
		~ & ~ & ~ & 1 & 10 & 50 & ~ & 1 & 10 & 50 & ~ & 1 & 10 & 50\\
    		\hline
		\multirow{2}*{0\%} & $\times$ & 94.51 & 80.90 & 95.09 & 98.19 & 85.84 & 41.00 & 71.80 & 87.31 & 83.93 & 57.50 & 83.52 & 93.45\\
		~ & $\checkmark$ & \pmb{95.91} & \pmb{94.88} & \pmb{99.11} & \pmb{99.72} & \pmb{94.85} &\pmb{90.83} & \pmb{98.29} & \pmb{99.08} & \pmb{89.83} & \pmb{94.05} & \pmb{98.72} & \pmb{99.57}\\
		\hdashline[2pt/2pt]
		\multirow{2}*{10\%} & $\times$ & 92.55 & 40.69 & 65.64 & 79.69 & 90.08 & 41.08 & 66.21 & 80.72 & \pmb{83.84} & 39.03 & 62.23 & 76.74\\
		~ & $\checkmark$ & \pmb{92.90} & \pmb{41.24} & \pmb{65.74} & \pmb{80.16} & \pmb{91.24} & \pmb{50.42} & \pmb{73.35} & \pmb{85.18} & 83.38 & \pmb{40.20} & \pmb{63.40} & \pmb{77.52}\\
		\hdashline[2pt/2pt]
		\multirow{2}*{30\%} & $\times$ & 82.18 & 10.93 & 26.52 & 42.02 & 77.20 & 10.80 & 25.61 & 45.56 & 70.35 & 4.72 & 15.05 & 29.47\\
		~ & $\checkmark$ & \pmb{84.85} & \pmb{12.87} & \pmb{29.39} & \pmb{45.58} & \pmb{82.85} & \pmb{12.19} & \pmb{29.37} & \pmb{46.53} & \pmb{71.07} & \pmb{5.65} & \pmb{17.40} & \pmb{32.90}\\
  		\hline
	\end{tabular}
	\caption{Controllable sketch synthesis (Mask~=~0\%) and sketch healing performance of SP-gra2seq by applying the clustering constraint over the inter-sketch patch embeddings or not.}
	\label{tab:ab_study_clustering}
\end{table*}

We also present the qualitative comparisons in Fig. \ref{fig:sketch_healing}. When a sketch is with masked key characteristics, e.g., the pig head with missing top part in the red bounding box, SP-gra2seq recognizes its category from the partial corrupted nose and successfully recreates a pig head. But the others fail. SP-gra2seq is powerful to restore the original sketch details during healing.

\subsection{Performance Gained from Synonymous Linkings}
Synonymous proximity activates the message communications between the sketch patches which are positioned in a long distance on the canvas by capturing their dynamic long-range dependencies as \citep{zhang2020dynamic}. This section verifies whether the performance gained by SP-gra2seq is from the connections between the synonymous patches or from the ones between the ``non-local'' patches. We randomly link the non-local patches by filling the adjacency matrix $\bm A_t$ in Eq. (\ref{eq:adj_mat}) with the values sampled from the uniform distribution $\mathcal{U}(0, 1)$, which activate the communications between the long-ranged patches without considering their synonymous relations. The results are in Table \ref{tab:ab_study_random}. The message passing by random affinities interrupts the model to focus on the local details in target patch. The performance drops especially when the sketches are lightly corrupted. It demonstrates that our performance gain is from the proposed synonymous proximity.

\subsection{The Impact of Clustering Constraint}
We cancel the clustering process in Eq. (\ref{eq:em_step}) and remove the second term from the objective in Eq. (\ref{eq:loss}). Table \ref{tab:ab_study_clustering} offers the quantitative results for controllable sketch synthesis and sketch healing, respectively. When training SP-gra2seq without the clustering constraint, the performance for both tasks reduces in most cases. It is because the synonymous patches are not encoded in compact clusters in the latent space. If a patch is covered with masks or is challenging to be recognized due to the drawing manner, the captured embedding could be far away from the cluster centroid, where the patch ought to be mapped closely. As a result, the patch embeddings may be unreliable and further reduce the sketch representing performance.

\section{Conclusions}

We have presented SP-gra2seq for learning efficient graphic sketch representations. Intra-sketch patches are linked with graph edges by synonymous proximity, which is learnable to adapt to the huge variation of sketch drawings. Accordingly, the message aggregation from the synonymous patches plays a role of denoising, leading to robust and reliable patch embeddings and accurate sketch representations for the downstream tasks. Moreover, we apply a clustering constraint over the inter-sketch patch embeddings jointly with the network training. Synonymous patches are encoded in compact clusters to yield accurate and robust computation of the synonymous proximity. Experiments on the tasks of controllable sketch synthesis and sketch healing demonstrate the effectiveness of our method.

\section{Acknowledgments}
This work was supported by the National Key R\&D Program of China (2018AAA0100700) of the Ministry of Science and Technology of China, and Shanghai Municipal Science and Technology Major Project (2021SHZDZX0102). Shikui Tu and Lei Xu are the corresponding authors.

\bibliography{ref_aaai_23}

\begin{thebibliography}{25}
\providecommand{\natexlab}[1]{#1}

\bibitem[{Beyer et~al.(1999)Beyer, Goldstein, Ramakrishnan, and
  Shaft}]{beyer1999nearest}
Beyer, K.; Goldstein, J.; Ramakrishnan, R.; and Shaft, U. 1999.
\newblock When is ¡°nearest neighbor¡± meaningful?
\newblock In \emph{International conference on database theory}, 217--235.
  Springer.

\bibitem[{Chen et~al.(2017)Chen, Tu, Yi, and Xu}]{chen2017sketch}
Chen, Y.; Tu, S.; Yi, Y.; and Xu, L. 2017.
\newblock Sketch-pix2seq: A model to generate sketches of multiple categories.
\newblock \emph{arXiv preprint arXiv:1709.04121}.

\bibitem[{Choi et~al.(2019)Choi, Cho, Song, and Yoon}]{choi2019sketchhelper}
Choi, J.; Cho, H.; Song, J.; and Yoon, S.~M. 2019.
\newblock Sketchhelper: Real-time stroke guidance for freehand sketch
  retrieval.
\newblock \emph{IEEE Transactions on Multimedia}, 21(8): 2083--2092.

\bibitem[{Ha and Eck(2018)}]{ha2017neural}
Ha, D.; and Eck, D. 2018.
\newblock A neural representation of sketch drawings.
\newblock In \emph{International Conference on Learning Representations}.

\bibitem[{Jang, Gu, and Poole(2016)}]{jang2016categorical}
Jang, E.; Gu, S.; and Poole, B. 2016.
\newblock Categorical reparameterization with Gumbel-softmax.
\newblock \emph{arXiv preprint arXiv:1611.01144}.

\bibitem[{Kipf and Welling(2016)}]{kipf2016semi}
Kipf, T.~N.; and Welling, M. 2016.
\newblock Semi-supervised classification with graph convolutional networks.
\newblock \emph{arXiv preprint arXiv:1609.02907}.

\bibitem[{Li et~al.(2022)Li, Jiang, Guan, Wang, and
  Thalmann}]{li2022multistage}
Li, H.; Jiang, X.; Guan, B.; Wang, R.; and Thalmann, N.~M. 2022.
\newblock Multistage spatio-temporal networks for robust sketch recognition.
\newblock \emph{IEEE Transactions on Image Processing}, 31: 2683--2694.

\bibitem[{Li et~al.(2021)Li, Jiang, Guarn, and Thalmann}]{li2021efficient}
Li, H.; Jiang, X.; Guarn, B.; and Thalmann, N.~M. 2021.
\newblock Efficient sketch recognition via compact spatial embedding graph
  neural networks.
\newblock In \emph{2021 IEEE International Conference on Multimedia and Expo},
  1--6.

\bibitem[{Lin et~al.(2020)Lin, Fu, Xue, and Jiang}]{lin2020sketch}
Lin, H.; Fu, Y.; Xue, X.; and Jiang, Y.-G. 2020.
\newblock Sketch-bert: Learning sketch bidirectional encoder representation
  from transformers by self-supervised learning of sketch gestalt.
\newblock In \emph{Proceedings of the IEEE/CVF Conference on Computer Vision
  and Pattern Recognition}, 6758--6767.

\bibitem[{Qi et~al.(2022{\natexlab{a}})Qi, Gryaditskaya, Xiang, and
  Song}]{qi2022one}
Qi, A.; Gryaditskaya, Y.; Xiang, T.; and Song, Y.-Z. 2022{\natexlab{a}}.
\newblock One sketch for all: One-shot personalized sketch segmentation.
\newblock \emph{IEEE Transactions on Image Processing}.

\bibitem[{Qi et~al.(2021)Qi, Su, Chowdhury, Li, and Song}]{qi2021sketchlattice}
Qi, Y.; Su, G.; Chowdhury, P.~N.; Li, M.; and Song, Y.-Z. 2021.
\newblock SketchLattice: Latticed representation for sketch manipulation.
\newblock In \emph{Proceedings of the IEEE/CVF International Conference on
  Computer Vision}, 953--961.

\bibitem[{Qi et~al.(2022{\natexlab{b}})Qi, Su, Wang, Yang, Pang, and
  Song}]{qi2022generative}
Qi, Y.; Su, G.; Wang, Q.; Yang, J.; Pang, K.; and Song, Y.-Z.
  2022{\natexlab{b}}.
\newblock Generative Sketch Healing.
\newblock \emph{International Journal of Computer Vision}, 1--16.

\bibitem[{Ribeiro et~al.(2020)Ribeiro, Bui, Collomosse, and
  Ponti}]{ribeiro2020sketchformer}
Ribeiro, L. S.~F.; Bui, T.; Collomosse, J.; and Ponti, M. 2020.
\newblock Sketchformer: Transformer-based representation for sketched
  structure.
\newblock In \emph{Proceedings of the IEEE/CVF Conference on Computer Vision
  and Pattern Recognition}, 14153--14162.

\bibitem[{Scarselli et~al.(2008)Scarselli, Gori, Tsoi, Hagenbuchner, and
  Monfardini}]{scarselli2008graph}
Scarselli, F.; Gori, M.; Tsoi, A.~C.; Hagenbuchner, M.; and Monfardini, G.
  2008.
\newblock The graph neural network model.
\newblock \emph{IEEE Transactions on Neural Networks}, 20(1): 61--80.

\bibitem[{Song et~al.(2018)Song, Pang, Song, Xiang, and
  Hospedales}]{song2018learning}
Song, J.; Pang, K.; Song, Y.-Z.; Xiang, T.; and Hospedales, T.~M. 2018.
\newblock Learning to sketch with shortcut cycle consistency.
\newblock In \emph{Proceedings of the IEEE Conference on Computer Vision and
  Pattern Recognition}, 801--810.

\bibitem[{Su et~al.(2020)Su, Qi, Pang, Yang, and Song}]{su2020sketchhealer}
Su, G.; Qi, Y.; Pang, K.; Yang, J.; and Song, Y.-Z. 2020.
\newblock Sketchhealer a graph-to-sequence network for recreating partial human
  sketches.
\newblock In \emph{Proceedings of The 31st British Machine Vision Virtual
  Conference}, 1--14. British Machine Vision Association.

\bibitem[{Tian et~al.(2021)Tian, Xu, Wang, Shen, and
  Liu}]{tian2021relationship}
Tian, J.; Xu, X.; Wang, Z.; Shen, F.; and Liu, X. 2021.
\newblock Relationship-preserving knowledge distillation for zero-shot sketch
  based image retrieval.
\newblock In \emph{Proceedings of the 29th ACM International Conference on
  Multimedia}, 5473--5481.

\bibitem[{Xu et~al.(2020)Xu, Huang, Yuan, Xiang, Hospedales, Song, and
  Wang}]{xu2020learning}
Xu, P.; Huang, Y.; Yuan, T.; Xiang, T.; Hospedales, T.~M.; Song, Y.-Z.; and
  Wang, L. 2020.
\newblock On learning semantic representations for million-scale free-hand
  sketches.
\newblock \emph{arXiv preprint arXiv:2007.04101}.

\bibitem[{Xu, Joshi, and Bresson(2021)}]{xu2021multigraph}
Xu, P.; Joshi, C.~K.; and Bresson, X. 2021.
\newblock Multigraph transformer for free-hand sketch recognition.
\newblock \emph{IEEE Transactions on Neural Networks and Learning Systems}.

\bibitem[{Yang et~al.(2020)Yang, Sain, Li, Qi, Zhang, and Song}]{yang2020s}
Yang, L.; Sain, A.; Li, L.; Qi, Y.; Zhang, H.; and Song, Y.-Z. 2020.
\newblock S3net: Graph representational network for sketch recognition.
\newblock In \emph{2020 IEEE International Conference on Multimedia and Expo},
  1--6. IEEE.

\bibitem[{Yang et~al.(2021)Yang, Zhuang, Fu, Wei, Zhou, and
  Zheng}]{yang2021sketchgnn}
Yang, L.; Zhuang, J.; Fu, H.; Wei, X.; Zhou, K.; and Zheng, Y. 2021.
\newblock SketchGNN: Semantic sketch segmentation with graph neural networks.
\newblock \emph{ACM Transactions on Graphics}, 40(3): 1--13.

\bibitem[{Yu et~al.(2017)Yu, Yang, Liu, Song, Xiang, and
  Hospedales}]{yu2017sketch}
Yu, Q.; Yang, Y.; Liu, F.; Song, Y.-Z.; Xiang, T.; and Hospedales, T.~M. 2017.
\newblock Sketch-a-net: A deep neural network that beats humans.
\newblock \emph{International Journal of Computer Vision}, 122(3): 411--425.

\bibitem[{Zang, Tu, and Xu(2021)}]{zang2021controllable}
Zang, S.; Tu, S.; and Xu, L. 2021.
\newblock Controllable stroke-based sketch synthesis from a self-organized
  latent space.
\newblock \emph{Neural Networks}, 137: 138--150.

\bibitem[{Zhang et~al.(2020{\natexlab{a}})Zhang, Xu, Arnab, and
  Torr}]{zhang2020dynamic}
Zhang, L.; Xu, D.; Arnab, A.; and Torr, P.~H. 2020{\natexlab{a}}.
\newblock Dynamic graph message passing networks.
\newblock In \emph{Proceedings of the IEEE/CVF Conference on Computer Vision
  and Pattern Recognition}, 3726--3735.

\bibitem[{Zhang et~al.(2020{\natexlab{b}})Zhang, Zhang, Feng, Zhang, and
  Fan}]{zhang2020zero}
Zhang, Z.; Zhang, Y.; Feng, R.; Zhang, T.; and Fan, W. 2020{\natexlab{b}}.
\newblock Zero-shot sketch-based image retrieval via graph convolution network.
\newblock In \emph{Proceedings of the AAAI Conference on Artificial
  Intelligence}, volume~34, 12943--12950.

\end{thebibliography}

\end{document}